\pdfoutput=1

\documentclass[11pt]{article}

\usepackage{naacl2021}

\usepackage{times}
\usepackage{latexsym}
\usepackage{lipsum}
\usepackage{hyperref}
\usepackage{verbatim}
\usepackage{graphicx}
\usepackage{arabtex}
\usepackage{utf8}
\usepackage{makecell}
\usepackage{booktabs}
\usepackage{fixltx2e}

\usepackage[T1]{fontenc}

\usepackage[utf8]{inputenc}

\usepackage{microtype}

\usepackage{cancel}
\usepackage{xpatch}
\usepackage[normalem]{ulem}

\title{AraStance: A Multi-Country and Multi-Domain Dataset of \\ Arabic Stance Detection for Fact Checking}

\author{\hspace{0.2cm}Tariq Alhindi,$^1$
        \hspace{0.2cm}Amal Alabdulkarim,$^2$
        \hspace{0.2cm}Ali Alshehri,$^3$\\
        \hspace{0.2cm}\textbf{Muhammad Abdul-Mageed,$^4$ and Preslav Nakov$^5$}\\     
  $^1$Department of Computer Science, Columbia University\\
  $^2$Georgia Institute of Technology,
  $^3$The State University of New York at Buffalo\\
  $^4$The University of British Columbia,
  $^5$Qatar Computing Research Institute, HBKU\\
  {\tt tariq@cs.columbia.edu},\hspace{0.1cm}
  {\tt amal@gatech.edu},\hspace{0.1cm}
  {\tt alimoham@buffalo.edu}\\
  {\tt muhammad.mageed@ubc.ca},\hspace{0.1cm}
  {\tt pnakov@hbku.edu.qa}
  }

\begin{document}
\maketitle
\setcode{utf8}
\begin{abstract}
With the continuing spread of misinformation and disinformation online, it is of increasing importance to develop combating mechanisms at scale in the form of automated systems that support multiple languages. One task of interest is claim veracity prediction, which can be addressed using stance detection with respect to relevant documents retrieved online. To this end, we present our new Arabic Stance Detection dataset (AraStance) of 4,063 claim--article pairs from a diverse set of sources comprising three fact-checking websites and one news website. AraStance covers false and true claims from multiple domains (e.g., politics, sports, health) and several Arab countries, and it is well-balanced between related and unrelated documents with respect to the claims. We benchmark AraStance, along with two other stance detection datasets, using a number of BERT-based models. Our best model achieves an accuracy of 85\% and a macro F1 score of 78\%, which leaves room for improvement and reflects the challenging nature of AraStance and the task of stance detection in general.
\end{list}
\end{abstract}

\section{Introduction} 
\label{sec:intro}

The proliferation of social media has made it possible for individuals and groups to share information quickly. While this is useful in many situations such as emergencies, where disaster management efforts can make use of shared information to allocate resources, this evolution can also be dangerous, e.g., when the news shared is not precise or is even intentionally misleading. Polarization in different communities further aggravates the problem, causing individuals and groups to believe and to disseminate information without necessarily verifying its veracity (misinformation) or even making up stories that support their world views (disinformation). These circumstances motivate a need to develop tools for detecting fake news online, including for a region with opposing forces and ongoing conflicts such as the Arab world. 

Our work here contributes to these efforts a new dataset and baseline results on it. In particular, we create a new dataset for stance detection of claims collected from a number of websites covering different domains such as politics, health, and economics. The websites cover several Arab countries, which enables wider applicability of our dataset. This compares favorably to previous work for Arabic stance detection such as the work of~\citet{baly-etal-2018-integrating}, who focused on a single country. We use the websites as our source to collect true and false claims, and we carefully crawl web articles related to these claims. Using the claim--article pairs, we then manually assign \textit{stance} labels to the articles. By stance we mean whether an article \textit{agrees, disagrees, discusses} a claim or it is just \textit{unrelated}. This allows us to exploit the resulting dataset to build models that automatically identify the stance with respect to a given claim, which is an important component of fact-checking and fake news detection systems. To develop these models, we resort to transfer learning by fine-tuning language models on our labeled dataset. We also benchmark our models on two existing datasets for Arabic stance detection. Finally, we make our dataset publicly available.\footnote{The data can be found at \url{http://github.com/Tariq60/arastance}.}

Our contributions can be summarized as follows:

\begin{enumerate}
    \item We release a new multi-domain, multi-country dataset labeled for both stance and veracity.
    \item We introduce a multi-query related document retrieval approach for claims from diverse topics in Arabic, resulting in a dataset with balanced label distributions across classes.  
    \item We compare our dataset to two other Arabic stance detection datasets using four BERT-based \cite{devlin-etal-2019-bert} models.
\end{enumerate}

\section{Related Work}
\label{sec:related_work}

Stance detection started as a standalone task, unrelated to fact-checking~\cite{kuccuk2020stance}. One type of stance models the relation (e.g.,~\emph{for, against, neutral}) of a text segment towards a \emph{topic}, usually a controversial one such as abortion or gun control \cite{mohammad2016semeval, abbott2016internet}. Another one models the relation (e.g.,~\emph{agree, disagree, discuss, unrelated}) between two pieces of text \cite{hardalov2021survey, ferreira2016emergent}. The latter definition is used in automatic fact-checking, fake news detection, and rumour verification \cite{vlachos2014fact}.

There are several English datasets that model fact-checking as a stance detection task on text from multiple genres such as Wikipedia \cite{thorne-etal-2018-fact}, news articles \cite{pomerleau2017fake, ferreira2016emergent}, and social media \cite{gorrell2019semeval,derczynski2017semeval}. Most related to our work here is the Fake News Challenge, or FNC, \cite{pomerleau2017fake}, which is built by randomly matching claim--article pairs from the Emergent dataset \cite{ferreira2016emergent}, which itself pairs 300 claims to 2,500 articles. In FNC, this pairing is done at random, and it yielded a large number of unrelated claim--article pairs. There are several approaches attempting to predict the stance on the FNC dataset using LSTMs, memory networks, and transformers \cite{hanselowski2018retrospective, conforti2018towards, mohtarami2018automatic, zhang2019stances, schiller2020stance, schutz2021automatic}.

There are two datasets for Arabic stance detection with respect to claims. The first one collected their false claims from a single political source \cite{baly-etal-2018-integrating}, while we cover three sources from multiple countries and topics. They retrieved relevant documents and annotated the claim--article pairs using the four labels listed earlier (i.e.,~\emph{agree, disagree, discuss, unrelated}). They also annotated ``rationales,'' which are segments in the articles where the stance is most strongly expressed. 
The other Arabic dataset by \citet{khouja-2020-stance} uses headlines from news sources and generated true and false claims by modifying the headlines. They used a three-class labeling scheme of stance by merging the {\it discuss} and the {\it unrelated} classes in one class called {\it other}. 

Our work is also related to detecting machine-generated and manipulated text~\cite{jawahar2020automatic,nagoudi2020machine}.

\begin{table*}
    \centering
    \scalebox{0.8}{
    \begin{tabular}{c|c}
    \toprule
         \textbf{Original Claim} & \textbf{Preprocessed Claim}\\
         \midrule
         \makecell{
          \<لا صحة لما يتم تداوله حول تسجيل>
         \\
            \<مكالمات ورسائل المقيمين>
            \\{\it What is being circulated about recording} \\ {\it residents' calls and messages is not true}}
          &\makecell{
            \<الحكومة تسجل مكالمات ورسائل المقيمين>
          \\{\it The government is recording calls and messages of residents}}
          \\
          \midrule
          
          \makecell{
            \<اشعة الكوزمو القادمة من كوكب المريخ اشاعة كاذبة من2008>
          \\{\it Cosmo rays coming from Mars} \\ {\it is a false rumor from 2008}
          }
          &
          \makecell{
            \<ناسا تحذر من اشعة الكوزمو الخطيرة القادمة من>
            \\
            \<كوكب المريخ  هذه الليلة من الساعة 12.30-3.30>
          \\{\it NASA warns of dangerous cosmic rays coming} \\ {\it from Mars tonight from 12.30-3.30}
          }
          \\
    \bottomrule
    \end{tabular}
    }
    \caption{Examples of false claims before and after preprocessing.}
    \label{tab:claim_preprocessing}
\end{table*}

\section{AraStance Construction}
\label{sec:data}

We constructed our AraStance dataset similarly to the way this was done for the English Fake News Challenge (FNC) dataset \cite{pomerleau2017fake} and for the Arabic dataset of \citet{baly-etal-2018-integrating}. Our dataset contains true and false claims, where each claim is paired with one or more documents. Each claim--article pair has a stance label: \emph{agree}, \emph{disagree}, \emph{discuss}, or \emph{unrelated}. Below, we decribe the three steps of building AraStance: (\emph{i})~claim collection and pre-processing, (\emph{ii})~relevant document retrieval, and (\emph{iii})~stance annotations.

\subsection{Claim Collection and Preprocessing}

We collected false claims from three fact-checking websites: {\sc Araanews\footnote{\url{http://araanews.ae}}, Dabegad\footnote{\url{http://dabegad.com}},} and {\sc Norumors\footnote{\url{http://norumors.net}}}, based in the UAE, Egypt, and Saudi Arabia, respectively. The claims were from 2012 to 2018 and covered multiple domains such as politics, sports, and health. As the three fact-checking websites only debunk false claims, we looked for another source for true claims: following \citet{baly-etal-2018-integrating}, we collected true claims from the Arabic website of {\sc Reuters\footnote{\url{http://ara.reuters.com}}}, assuming that their content was trustworthy. We added topic and date restrictions when collecting the true claims in order to make sure they were similar to the false claims. Moreover, in order to ensure the true claims were from the same topics as the false ones, we used a subset of the false claims as seeds to retrieve true claims that were within three months of the seed false claims, and we ranked them by TF.IDF, similarity to the seeds.
We kept a maximum of ten true claims per seed false claim. For all claims, we removed the ones that contained no-text and/or were multimedia-centric. Moreover, we manually modified the false claims by removing phrases like ``\emph{It is not true that}'', ``\emph{A debunked rumor about}'', or ``\emph{The reality of}'', which are often used by fact-checking websites. This sometimes required us to add a noun at the beginning of the claim based on the text of the target articles, or to make some grammatical edits. We show examples of two false claims before and after preprocessing in Table~\ref{tab:claim_preprocessing}. Note that the headlines we retrieved from {\sc reuters} were already phrased as claims, and thus we did not have to edit them in any way.

\subsection{Document Retrieval}
For each claim, we retrieved relevant documents using multiple queries and the Google Search API. It was harder to find relevant documents for the false claims by passing their preprocessed version as queries because of their nature, locality, and diversity. For some false claims, there were extra clauses and modifiers that restricted the search results significantly as shown in the examples below:

\begin{enumerate}
\item 
\<طفلة تظهر بنصف جسم بشري، \\ النصف الاخر ثعبان> 
\\
{\it A female child with half a human body, and the other half is a snake}
\item 
\<يتم تنظيف الرئتين عند المدخنين من خلال \\شم بخار اللبن و الماء لمدة عشرة أيام >
\\
{\it Lungs of smokers are cleaned by smelling the steam of milk and water for ten days}
\\
\end{enumerate}

\noindent To remedy this, we boosted the quality of the retrieved documents by restricting the date range to two months before and after the date of the claim, prepending named entities and removing extra clauses using parse trees. In order to emphasize the presence of the main entity(s) in the claim, we extracted named entities using the Arabic NER corpus by \citet{benajiba2007anersys} and Stanford's CoreNLP Arabic NER tagger \cite{manning2014stanford}. We further used Stanford's CoreNLP Arabic parser to extract the first verb phrase (VP) and all its preceeding tokens in the claim, as this has been shown to improve document retrieval results for claim verification, especially for lengthy claims \cite{chakrabarty-etal-2018-robust}. For the two examples shown above, we would keep the claims until the comma for the first example and the word \emph{and} for the second one, and we would consider those as the queries.

For each false claim, we searched for relevant documents using the following five queries: (\emph{i})~the manually preprocessed claim as is, (\emph{ii})~the preprocessed claim with date restriction, (\emph{iii})~the preprocessed claim with named entities and date restriction, (\emph{iv})~the first VP and all preceding tokens with date restriction, and lastly (\emph{v})~the first VP and all preceding tokens with named entities and date restriction. For the true claims, due to wider coverage that led to easier retrieval, we only ran two queries, using the claim with and without date restriction.

\begin{table*}[tbh]
    \centering
    \scalebox{0.78}{
    \begin{tabular}{c|c|c|c|c}
    \toprule
         \textbf{Claim} & \textbf{Document Title} & \textbf{A1} & \textbf{A2} & \textbf{A3} \\
         \midrule
         \makecell{\footnotesize{
            \<اللحوم والدواجن المستوردة من البرازيل فاسدة>
            }\\{\it Meat and poultry imported from Brazil are rotten}}
          &\makecell{\footnotesize{
            \<فضيحة البرازيل: اللحوم الفاسدة تهدد الانتعاش الاقتصادي>
          }\\{\it Brazil scandal: Rotten meat threatens economic recovery}}
          & D & A & D
          \\\midrule
          
          \makecell{\footnotesize{
            \<تشارك مصر في عملية انقاذ أطفال الكهف فى تايلاند>
          }\\{\it Egypt participates in the rescue operation}\\ {\it of cave children in Thailand}}
          &\makecell{\footnotesize{
            \<الإنقاذ يبدأون عملية لإخراج صبية من داخل كهف في تايلاند>
          }\\{\it Rescuers begin operation to remove the} \\ {\it teenage boys from the cave in Thailand}}
          & D & U & U
          \\
    \bottomrule
    \end{tabular}}
    \caption{Disagreement between the annotators on the {\it discuss (D)} label with the {\it agree (A)} (first example) and the {\it unrelated (U)} labels (second example).}
    \label{tab:discuss}
\end{table*}

We combined the results from all queries, and we kept a maximum of ten documents per claim. If the retrieved documents exceeded this limit, we only kept documents from news sources\footnote{We used Google News as a reference of news sources for the three countries of the fact-checking websites:
\url{https://news.google.com/?hl=ar&gl=X&ceid=X\%3Aar}, where {\bf X} is {\bf AE, EG,} or {\bf SA}, standing for UAE, Egypt and Saudi Arabia, respectively.}, or from sources used in previous work on Arabic stance detection \cite{baly-etal-2018-integrating, khouja-2020-stance}. If we still had more than ten documents after filtering by source, we ranked the documents by their TF.IDF similarity with the claim, and we kept the top ten documents. We limited the number of documents to ten per claim in order to avoid having claims with very high numbers of documents and others with only one or two documents. Ultimately, this helped us keep the dataset balaced in terms of both sources and topics.

\subsection{Stance Annotation}
We set up the annotation task as follows: given a claim--article pair, what is the stance of the document towards the claim? The stance was to be annotated using one of the following labels: {\it agree, disagree, discuss}, or {\it unrelated}, which were also used in previous work \cite{pomerleau2017fake, baly-etal-2018-integrating}. 

\noindent We explained the labels to annotators as follows:

\begin{itemize}
  \item {\it agree:} the document agrees with the main claim in the statement clearly and explicitly;
  \item {\it disagree:} the document disagrees with main claim in the statement clearly and explicitly;
  \item {\it discuss:} the document discusses the same event without taking a position towards its validity;
  \item {\it unrelated:} the document talks about a different event, regardless of how similar the two events might be.
\end{itemize}

Our annotators were three graduate students in computer science and linguistics, all native speakers of Arabic. We adopted guidelines similar to the ones introduced by \newcite{baly-etal-2018-integrating}. First, we conducted a pilot annotation round on 315 claim--article pairs, where each pair was annotated by all annotators. The annotators agreed on the same label for 220 out of the 315 pairs (70\% of the pairs), while for 89 pairs (28\%) there were two annotators agreeing on the label, and for the remaining 6 pairs (2\% of the pairs) there was a three-way disagreement. The main disagreements between the annotators were related to the {\it discuss} label, which was confused with either {\it agree} or {\it unrelated}. 

We show two examples in Table \ref{tab:discuss} where the annotators labeled the example on the top of the table as {\it discuss} and {\it agree}. The two annotators that labeled this example as {\it discuss} justified their choice by arguing that the document only mentioned the claims without agreeing or disagreeing and mainly analyzed the impact of rotten meat on Brazil's economy in great detail. The example in the bottom of the table was labeled by one annotator as {\it discuss} and by two annotators as {\it unrelated}. The annotators who labeled it as {\it unrelated} argued that there was no mention of Egypt's involvement in the rescue efforts, while the annotator who labeled the pair as {\it discuss} maintained that the document discussed the same event of children trapped in the cave. 

These disagreements were resolved through discussions between the annotators, which involved refining the guidelines to label a pair as {\it discuss} if it only talks about the exact same event of the claim without taking any clear position. The annotators were also asked not to take into consideration any other factors, e.g.,~the date of article, its publisher, or its veracity.

For the rest of the data, each claim--article pair was annotated by two annotators, where the differences were resolved by the third annotator. This is very similar to labeling all pairs by three annotators with majority voting, but with less labor requirements. We measured the inter-annotator agreement (IAA) using Fleiss kappa, which accounts for multiple annotators \cite{fleiss1973equivalence}, obtaining an IAA of 0.67, which corresponds to substantial agreement.

\begin{table*}
    \centering
    \begin{tabular}{l|l r r|r r r r}
    \toprule
    & & & & \multicolumn{4}{|c}{\textbf{Stance}}\\
         \textbf{Source} & \textbf{Veracity} & \textbf{Claims} & \textbf{Articles} & \textbf{Agree} & \textbf{Disagree} & \textbf{Discuss} & \textbf{Unrelated}\\
    \midrule
    {\sc Aranews} &False &170 &518      &80 &82 &51 &305 \\
    {\sc Dabegaed} &False &278 &1,413     &225 &249 &143 &796  \\
    {\sc Norumers} &False &158 &490      &26 &103 &32 &329  \\
    {\sc Reuters} &True &304 &1,642      &691 &15 &161 &775  \\
    \hline
    Total &-- &910 &4,063      &1,022 &449 &387 &2,205  \\
    \bottomrule
    \end{tabular}
    \caption{Statistics about the number of claims, articles and claim--article pairs and the distribution of their stances for each source.}
    \label{tab:data}
\end{table*}

\begin{table}
    \centering
    \begin{tabular}{l|r r r}
    \toprule
    \textbf{Label} & \textbf{Train} &\textbf{Dev} &\textbf{Test}\\
    \midrule
         Agree &739 &129 &154 \\
         Disagree &309 &76 &64\\
         Discuss &247 &70 &70\\ 
         Unrelated &1,553 &294 &358\\
    \midrule
         Total &2,848 &569 &646\\
    \bottomrule
    \end{tabular}
    \caption{Statistics about the claim--article pairs with stances in the training, development and test sets.}
    \label{tab:stance}
\end{table}

\subsection{Statistics About the Final Dataset}
Table \ref{tab:data} shows the number of claims and articles for each website with their veracity label (by-publisher) and final stance annotations. The distribution of the four stance classes in training, development, and test is shown in Table~\ref{tab:stance}. After selecting the gold annotations, we discarded all claims that had all of their retrieved documents labeled as {\it unrelated}, aiming to reduce the imbalance with respect to the {\it unrelated} class, and we only focused on claims with related documents, which can be seen as a proxy for check-worthiness. 
We ended up with a total of 4,063 claim--articles pairs based on 910 claims: 606 false and 304 true. The dataset is imbalanced towards the false claims, but as our main task is stance detection rather than claim veracity, we aimed at having a balanced distribution for the four stance labels. As shown in Table~\ref{tab:stance}, around half of the labels are from the {\it unrelated} class, but it is common for stance detection datasets to have higher proportion of this class \cite{pomerleau2017fake, baly-etal-2018-integrating}. 

There are various approaches that can mitigate the impact of the class imbalance caused by the {\it unrelated} class. These are related to (\emph{i})~task setup, (\emph{ii})~modeling, and (\emph{iii})~evaluation. 

First, the task can be approached differently by only doing stance detection on the three related classes \cite{conforti2018towards}, or by merging the {\it discuss} and the {\it unrelated} classes into one class, e.g.,~called {\it neutral} or {\it other} \cite{khouja-2020-stance}.

Second, it is possible to keep all classes, but to train a two-step model: first to predict related vs. unrelated, and then, if the example is judged to be related, to predict the stance for the three related classes only \cite{zhang2019stances}. 

Third, one could adopt an evaluation measure that rewards models that make correct predictions for the related classes more than for the \emph{unrelated} class. Such a measure was adopted by the Fake News Challenge \cite{pomerleau2017fake}. However, such measures have to be used very carefully, as they might be exploited. For example, it was shown that the FNC measure can be exploited by random prediction from the related classes and never from the {\it unrelated} class, which has a lower reward under the FNC evaluation measure \cite{hanselowski2018retrospective}. We leave such considerations about the impact of class imbalance to future work.

\section{Experimental Setup}
\label{sec:setup}

\subsection{External Datasets}

We experimented with a number of BERT-based models, pre-trained on Arabic or on multilingual data, which we fine-tuned and applied to our dataset, as well as to the following two Arabic stance detection datasets for comparison purposes:
\begin{itemize}
    \item \textbf{\citet{baly-etal-2018-integrating} Dataset.} This dataset has 1,842 claim--article pairs for training (278~{\it agree}, 37~{\it disagree}, 266 {\it discuss}, and 1,261 {\it unrelated}), 587 for development (86~{\it agree}, 25~{\it disagree}, 73 {\it discuss}, and 403~{\it unrelated}), and 613 for testing (110 {\it agree}, 25~{\it disagree}, 70 {\it discuss}, and 408 {\it unrelated}).
    
    \item \textbf{\citet{khouja-2020-stance} Dataset.} This dataset has 2,652 claim--article pairs for training (903~{\it agree}, 1,686~{\it disagree}, and 63~{\it other}), 755 for development (268~{\it agree}, 471~{\it disagree}, and 16~{\it other}) and 379 for testing (130~{\it agree}, 242~{\it disagree}, and 7~{\it other}).
\end{itemize}

The dataset by \citet{baly-etal-2018-integrating} has 203 true claims from {\sc Reuters} and 219 false claims from the Syrian fact-checking website {\sc Verify-Sy},\footnote{\url{http://www.verify-sy.com/}} which focuses on debunking claims about the Syrian civil war. Thus, the dataset contains claims that focus primarily on war and politics. They retrieved the articles and performed manual annotation of claim--article pairs for stance, following a procedure  that is very close to the one we used for AraStance. Moreover, their dataset has annotations of rationales, which give the reason for selecting an {\it agree} or a {\it disagree} label. The dataset has a total of about 3,000 claim--article pairs, 2,000 of which are from the {\it unrelated} class. The dataset comes with a split into five folds of roughly equal sizes. We use folds 1-3 for training, fold 4 for development, and fold 5 for testing.

The dataset by \citet{khouja-2020-stance} is based on sampling a subset of news titles from the Arabic News Text (ANT) corpus \cite{chouigui2017ant}, and then making true and false alterations of these titles using crowd-sourcing. The stance detection task is then defined between pairs of original news titles and their respective true/false alterations. This essentially maps to detecting paraphrases for true alterations (stance labeled as {\it agree}) and contradictions for false ones (stance labeled as {\it disagree}). They further have a third stance label, {\it other}, which is introduced by pairing the alterations with other news titles that have high TF.IDF similarity with the news title originally paired with the alteration. 
Overall, \citet{khouja-2020-stance}'s dataset is based on \emph{synthetic statements} that are paired with \emph{news titles}. This is quite different from AraStance and the dataset of \citet{baly-etal-2018-integrating}, which have \emph{naturally occurring claims} that are paired with \emph{full news articles}. Moreover, as both AraStance and \citet{baly-etal-2018-integrating}'s datasets have naturally occurring data from the web, they both exhibit certain level of noise and irregularities, e.g.,~some very long documents, words/characters in other languages such as English, etc. Such a noise is minimal in \citet{khouja-2020-stance}'s dataset, which is a third differentiating factor compared to the other two datasets. Nevertheless, we include \citet{khouja-2020-stance}'s dataset in our experiments in order to empirically test the impact of these differences.

\subsection{Models}
We fine-tuned the following four models for each of the three Arabic datasets:
\begin{enumerate}
    \item Multilingual BERT (mBERT), base size, which is trained on the Wikipedias of 100 different languages, including Arabic \cite{devlin-etal-2019-bert}.
    \item ArabicBERT, base size, which is trained on 8.2 billion tokens from the OSCAR corpus\footnote{\url{http://oscar-corpus.com}} as well as on the Arabic Wikipedia \cite{safaya-etal-2020-kuisail}.
    \item ARBERT, which is trained on 6.2 billion tokens of mostly Modern Standard Arabic text \cite{mageed2020marbert}.
    \item MARBERT, which is trained on one billion Arabic tweets, which in turn use both Modern Standard Arabic and Dialectal Arabic \cite{mageed2020marbert}.
\end{enumerate}

The four models are comparable in size, all having a \textit{base} architecture, but with varying vocabulary sizes. More information about the different models can be found in the original publications about them. We fine-tuned each of them for a maximum of 25 epochs with an early stopping patience value of 5, a maximum sequence length of 512, a batch size of 16, and a learning rate of 2e-5.

\begin{table*}[t]
    \centering
    \scalebox{.85}{
    \begin{tabular}{l||c c c c|c c||c c c|c c||c c c c|c c}
         \toprule
         \textbf{Model} & \multicolumn{6}{c||}{\textbf{\citet{baly-etal-2018-integrating} Dataset}} & \multicolumn{5}{c||}{\textbf{\citet{khouja-2020-stance} Dataset}} & \multicolumn{6}{c}{\textbf{AraStance}}  \\ 
         & \textbf{A} & \textbf{D} & \textbf{Ds} & \textbf{U} & \textbf{Acc} & \textbf{F1} & \textbf{A} & \textbf{D} & \textbf{O} & \textbf{Acc} & \textbf{F1} &\textbf{A} & \textbf{D} & \textbf{Ds} & \textbf{U} & \textbf{Acc} & \textbf{F1}\\
         \midrule
         mBERT 
         &{\bf .63} &0 &.11 &{\bf .84}    &{\bf .73} &.40
         &.74 &.84 &.76      &.81 &.78
         &.81 &.68 &.58 &.92    &.82 &.75 \\
         ArabicBERT
         &.58 & {\bf.14} &.24 &.82  &.69 &.45
         &.74 &.86 &.84      &.82 &.81
         &{\bf.85} &.75 &.56 &.92    &.84 &.77 \\
         ARBERT 
         &.56 & {\bf.14} &{\bf .30} &.83  &.70 & {\bf.46}
         &{\bf .81} &{\bf .89} &{\bf .87}      &{\bf .86} &{\bf .86}
         &{\bf.85} & {\bf.82} & {\bf.60} & {\bf.93}     & {\bf.86} & {\bf.80} \\
         MARBERT 
         &.44 & {\bf.14} &.23 &.78  &.62 &.40
         &.80 &.88 &.79      &.85 &.82
         &{\bf.85} &.80 &.53 &.89    &.84 &.77 \\
         \bottomrule
    \end{tabular}}
    \caption{Results on the development set for the three Arabic Stance Detection datasets. Shown are the F1-scores for each class ({\bf A}: Agree, {\bf D}: Disagree, {\bf Ds}: Discuss, {\bf U}: Unrelated, {\bf O}: Other), as well as the overall Accuracy ({\bf Acc}), and the Macro-Average F1 score (Macro-F1).}
    \label{tab:results_dev}
\end{table*}

\begin{table*}[t]
    \centering
    \scalebox{.85}{
    \begin{tabular}{l||c c c c|c c||c c c|c c||c c c c|c c}
         \toprule
         \textbf{Model} & \multicolumn{6}{c||}{\textbf{\citet{baly-etal-2018-integrating} Dataset}} & \multicolumn{5}{c||}{\textbf{\citet{khouja-2020-stance} Dataset}} & \multicolumn{6}{c}{\textbf{AraStance}}  \\ 
         & \textbf{A} & \textbf{D} & \textbf{Ds} & \textbf{U} & \textbf{Acc} & \textbf{F1} & \textbf{A} & \textbf{D} & \textbf{O} & \textbf{Acc} & \textbf{F1} &\textbf{A} & \textbf{D} & \textbf{Ds} & \textbf{U} & \textbf{Acc} & \textbf{F1}\\
         \midrule
         mBERT 
         &.64 &0 &.12 &{\bf .85}    &{\bf .73} &.40
         &.67 &.81 &.86      &.76 &.78
         &.83 &.77 &.51 &.93    & {\bf.85} &.76 \\
         ArabicBERT
         &{\bf .66} &{\bf .35} &{\bf .27} &.80  &.67 &{\bf .52}
         &.72 & .85 &.71      &.81 &.76
         &.84 &.74 &.52 & {\bf.94 }   & {\bf.85} &.76 \\
         ARBERT 
         &.65 &.29 &{\bf .27} &.81  &.68 &.51
         &{\bf .80} &{\bf .89} &{\bf 1.0}      &{\bf .86} &{\bf .90}
         & .85 & {\bf.78} &{\bf .55} &.92     & {\bf.85} & {\bf.78} \\
         MARBERT 
         &.51 &0 &.25 &.77  &.60 &.38
         &.78 &.88 &.92      &.84 &.86
         & {\bf.86} &.72 &.41 &.90    &.84 &.72 \\
         \bottomrule
    \end{tabular}}
    \caption{Results on the test set for the three Arabic Stance Detection datasets. Shown are the F1-scores for each class ({\bf A}: Agree, {\bf D}: Disagree, {\bf Ds}: Discuss, {\bf U}: Unrelated, {\bf O}: Other), as well as the overall Accuracy ({\bf Acc}), and the Macro-Average F1 score (Macro-F1).}
    \label{tab:results_test}
\end{table*}

\section{Results}
\label{sec:Results}

The evaluation results are shown in Tables \ref{tab:results_dev} and \ref{tab:results_test} for the development and for the test sets, respectively. We use accuracy and macro-F1 to account for the different class distributions; we also report per-class F1 scores. Note that \citet{khouja-2020-stance} uses three labels rather than four, merging {\it discuss} and {\it unrelated} into {\it other}. Their label distribution has a majority of {\it disagree}, followed by {\it agree}, and very few instances of {\it other}, which is different from our dataset and from \citet{baly-etal-2018-integrating}'s. 

We can see that ARBERT yields the best overall and per-class performance on dev for the \citet{khouja-2020-stance} dataset and AraStance. It also generalizes very well to the test sets, where it even achieved a higher macro-F1 score for the \citet{khouja-2020-stance} dataset. The performance of the other three models (mBERT, ArabicBERT, and MARBERT) drops slightly on the test set compared to dev for both AraStance and the \citet{khouja-2020-stance} dataset. This might be due to ARBERT being pre-trained on more suitable data, which includes Books, Gigaword and Common Crawl data primarily from MSA, but also a small amount of Egyptian Arabic. Since half of our data comes from an Egyptian website ({\sc Dabegad}), this could be helpful. Indeed, while ArabicBERT is pretrained on slightly more data than ARBERT, it was almost exclusively pretrained on MSA, without dialectal data, and AraStance it performs worse.

About the other models: The datasets on which ArabicBERT was trained have duplicates, which could explain the model being outperformed. For MARBERT, it is pretrained on tweets that have both MSA and dialectal Arabic. MARBERT's data come from social media, which is different from the news articles or titles from which all the experimental downstream three datasets are derived. Also, it seems that ARBERT and MARBERT are better than the other two models at predicting the stance between a pair of sentences, as it is the case with the \citet{khouja-2020-stance} dataset. 

This could be due to the diversity of their pretraining data, which improves the model's ability to capture inter-sentence relations such as paraphrases and contradictions. Another factor that could explain ARBERT's better performace compared to MARBERT is that the latter is trained with a masking objective only, while ARBERT is trained with both a masking objective \textit{and} a next sentence prediction objective. The use of the latter objective by ARBERT could explain its ability to capture information in our claim--stance pairs, although these pairs are different from other types of pairs such as in the question and answer task, where the pair occurs in an extended piece of text.

On the other hand, there is no consistently best model for the \citet{baly-etal-2018-integrating} dataset. This could be due to a number of reasons. First, that dataset has a severe class imbalance, as we have explained in Section~\ref{sec:setup}. Second, the dataset (especially the false claims) is derived from one particular domain, i.e.,~the Syrian war, which might not be well represented in the pretraining data. Therefore, additional modeling considerations such as adaptive pretraining on a relevant unlabelled corpus before fine-tuning on the target labeled data could help. 

Surprisingly, ArabicBERT and ARBERT perform much better on the test set than on the development set of the \citet{baly-etal-2018-integrating} dataset for the {\it disagree} class, which has the lowest frequency: from 0.14 F1 to 0.29--0.35 F1. 

Since the number of {\it disagree} instances is very low (25 documents for 10--12 unique claims), it is possible that the claims in the test set happen to be more similar to the ones in the training data than it is for development. This is plausible because we did our train-dev-test split based on the five-folds prepared by the authors as explained in Section~\ref{sec:setup}. It is worth noting that the multilingual model (mBERT) has the highest overall accuracy and F1 score for the {\it unrelated} class of the \citet{baly-etal-2018-integrating} dataset. Multilingual text representations such as mBERT might over-predict from the majority class, and thus would perform poorly on the two low-frequency classes; indeed, mBERT has an F1-score of 0 for {\it disagree}, and no more than 0.12 for {\it discuss} on development and testing.

Finally, we observe very high performance for all models for the {\it unrelated} class of AraStance. This could be an indication of strong signals that differentiate the \emph{related} and the \emph{unrelated} classes, whereas the {\it discuss} class is the most challenging one in AraStance, due to its strong resemblance to {\it agree} in some examples such as the one shown in Table \ref{tab:discuss}. This indicates that all models offer an area for improvement, where a single classifier can excel for both frequent and infrequent classes for the stance detection within and across datasets. We leave further experimentation, including with models developed for FNC and the \citet{baly-etal-2018-integrating} dataset, for future work.

\section{Conclusion and Future Work}
\label{sec:conc}

We presented AraStance, a new multi-topic Arabic stance detection dataset with claims extracted from multiple fact-checking sources across three countries and one news source. We discussed the process of data collection and approaches to overcome challenges in related document retrieval for claims with low online presence, e.g.,~due to topic or country specificity. We further experimented with four BERT-based models and two additional Arabic stance detection datasets.

In future work, we want to further investigate the differences between the three Arabic stance detection datasets and to make attempts to mitigate the impact of class imbalance, e.g.,~by training with weighted loss, by upsampling or downsampling the classes, etc. We further want to examine the {\it discuss} class across datasets and to compare the choice of annotation scheme ---three-way vs. four-way--- on this task. Moreover, we plan to enrich AraStance by collecting more true claims from other websites, thus creating a dataset that would be more evenly distributed across the claim veracity labels. Furthermore, we would like to investigate approaches for improving stance detection by extracting the parts of the documents that contain the main stance rather than truncating the documents after the first 512 tokens. Finally, we plan to experiment with cross-domain \cite{hardalov2021crossdomain} and cross-language approaches \cite{mohtarami-etal-2019-contrastive}.

\section*{Acknowledgements}\label{sec:acknow}
Tariq Alhindi is supported by the KACST Graduate Studies Scholarship. Muhammad Abdul-Mageed gratefully acknowledges support from the Social Sciences and Humanities Research Council of Canada through an Insight Grant on contextual misinformation detection for Arabic social media, the Natural Sciences and Engineering Research Council of Canada, Canadian Foundation for Innovation, Compute Canada, and UBC ARC-Sockeye.\footnote{\url{https://doi.org/10.14288/SOCKEYE}.} Preslav Nakov is supported by the Tanbih mega-project,
which is developed at the Qatar Computing Research Institute, HBKU, and aims to limit the impact of ``fake news,'' propaganda, and media bias by making users aware of what they are reading. We would also like to thank Firas Sabbah for his help in collecting the false and true claims, and the anonymous reviewers for their helpful feedback.

\bibliography{anthology,custom}
\bibliographystyle{acl_natbib}

\end{document}